\documentclass[10pt,twocolumn,letterpaper]{article}

\usepackage{iccv}              

\usepackage[accsupp]{axessibility}  
\usepackage{times}
\usepackage{epsfig}
\usepackage{graphicx}
\usepackage{amsmath}
\usepackage{amssymb}
\usepackage{bbding}

\usepackage{csquotes}

\usepackage{multirow}
\usepackage{graphicx}
\usepackage{booktabs}
\usepackage{bbm}

\usepackage{color, colortbl}

\definecolor{mygray}{gray}{0.95}

\usepackage{array}
\usepackage{tabularx}
\usepackage{makecell}
\usepackage{algorithm}
\usepackage{algorithmic}
\usepackage{multirow}
\usepackage{xcolor}

\usepackage{color, colortbl}
\definecolor{Gray}{gray}{0.9}

\usepackage{pifont}
\newcommand\blfootnote[1]{%
  \begingroup
  \renewcommand\thefootnote{}\footnote{#1}%
  \addtocounter{footnote}{-1}%
  \endgroup
}

\definecolor{iccvblue}{rgb}{0.21,0.49,0.74}
\usepackage[pagebackref,breaklinks,colorlinks,allcolors=iccvblue]{hyperref}

\def\paperID{}
\def\confName{}
\def\confYear{}

\title{
R1-VL: Learning to Reason with Multimodal Large Language Models via Step-wise Group Relative Policy Optimization
}

\author{Jingyi Zhang \ \ Jiaxing Huang$^{ \text{\Envelope}}$  \  Huanjin Yao \ \  Shunyu Liu  \ \  Xikun Zhang  \ \    Shijian Lu \ \  Dacheng Tao
\\Nanyang Technological University, Singapore
}

\begin{document}

\maketitle

\begin{abstract}

Recent studies generally enhance MLLMs' reasoning capabilities via supervised fine-tuning on high-quality chain-of-thought reasoning data, which often leads models to merely imitate successful reasoning paths without understanding what the wrong reasoning paths are.
In this work, we aim to enhance the MLLMs' reasoning ability beyond passively imitating positive reasoning paths. 
To this end, we design Step-wise Group Relative Policy Optimization (StepGRPO), a new online reinforcement learning framework that enables MLLMs to self-improve reasoning ability via simple, effective and dense step-wise rewarding.
Specifically, StepGRPO introduces two novel rule-based reasoning rewards: 
Step-wise Reasoning Accuracy Reward (StepRAR) and Step-wise Reasoning Validity Reward (StepRVR).
StepRAR rewards the reasoning paths that contain necessary intermediate reasoning steps via a soft key-step matching technique, while StepRAR rewards reasoning paths that follow a well-structured and logically consistent reasoning process through a reasoning completeness and logic evaluation strategy.
With the proposed StepGRPO, we introduce R1-VL, a series of MLLMs with outstanding capabilities in step-by-step reasoning. 
Extensive experiments over 8 benchmarks demonstrate the superiority of our methods. Code is available at \href{https://github.com/jingyi0000/R1-VL}{link}.

\end{abstract}

\blfootnote{
Correspondence to: Jiaxing Huang \{jiaxing.huang@ntu.edu.sg\}.
} 

\section{Introduction}\label{introduction}

Multimodal large language models (MLLMs) have achieved significant progress in vision-language understanding~\cite{gpt4o,claude_3.5_sonnet,Cambrian-1,mm1.5,idefics3,internvl2,deepseek-vl2,llava}.
Recent efforts generally enhance MLLMs' reasoning capabilities by employing supervised fine-tuning (SFT) on high-quality chain-of-thought (CoT) reasoning data generated by powerful models (e.g., GPT4)~\cite{llava-cot,llava-reasoner,thawakar2025llamav,yao2024mulberry}. 
For example, Mulberry~\cite{yao2024mulberry} introduces CoMCTS, which utilizes multiple models to collectively search and identify effective reasoning paths, followed by SFT on the collected reasoning data.
However, SFT approaches focus solely on positive reasoning paths (i.e., those leading to correct answers), while the negative reasoning paths are largely neglected. This limitation may cause the model to merely imitate successful reasoning paths without understanding what the flawed and wrong reasoning paths are.

\begin{figure}[t]
\centering
\includegraphics[width=0.9\linewidth]{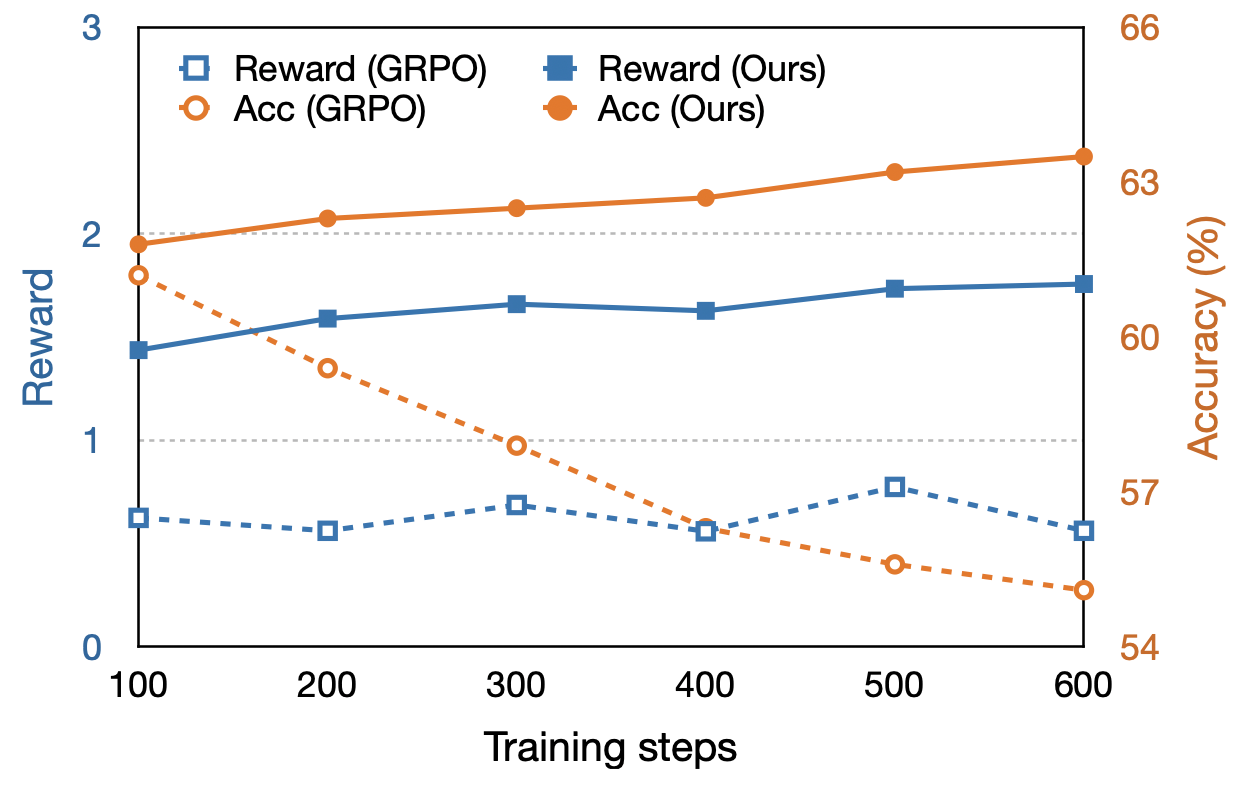}
\caption{
For MLLMs, online reinforcement learning with outcome-level reward, like in Deepseek-R1's GRPO~\cite{shao2024deepseekmath}, often suffers from sparse reward issues, where only a few reasoning paths can receive positive/high rewards during training, ultimately leading to poor exploration efficiency and unstable learning process. 
To tackle this, we propose a novel online reinforcement learning framework that incorporates step-wise reasoning rewards in addition to outcome-level rewards, encouraging MLLMs to iteratively refine their reasoning with dense rewards and resulting in a more stable training process and improved reasoning capability.
The experiments are conducted on Qwen2-VL-7b over MathVista.
}
\label{fig:1}
\end{figure}

In this work, we aim to enhance the MLLMs' reasoning ability beyond passively imitating positive reasoning paths.
Recent advancements in NLP, such as Deepseek-R1~\cite{guo2025deepseek} and Kimi-K1.5~\cite{team2025kimi}, have shown great potential in incentivizing the reasoning capability of LLMs via actively self-exploring.
The core design of these advances (e.g., GRPO in Deepseek-R1) lies in online reinforcement learning without the need for reward models, which encourages an LLM to generate a group of reasoning paths and iteratively refine its reasoning process by rewarding the generated reasoning paths based on a rule-based reward function.
Typically, an outcome-level reward strategy is used: reasoning paths leading to correct answers receive higher rewards, while those leading to incorrect answers receive lower ones.

An intuitive idea is to directly apply these simple and effective LLM online reinforcement learning methods for MLLMs. 
However, relying solely on outcome-level rewards, like in Deepseek-R1's GRPO, often suffers from sparse reward issues on MLLM reasoning learning, resulting in suboptimal performance. 
Specifically, most MLLMs, especially smaller ones, exhibit very limited capability in long-chain reasoning accuracy and validity, whereas only a few MLLM-generated reasoning paths can receive positive/high rewards. This lack of positive reward signals reduces exploration efficiency and leads to an unstable learning process, as illustrated in Fig.~\ref{fig:1}.

We propose to tackle this sparse reward issue by introducing dense step-wise reasoning rewards in addition to sparse outcome-level rewards.
To this end, we design Step-wise Group Relative Policy Optimization (StepGRPO), a new online reinforcement learning framework that enables MLLMs to self-improve reasoning ability via simple, effective and dense step-wise rewarding while using no additional process reward models.
Specifically, StepGRPO introduces two novel rule-based reasoning reward mechanisms: Step-wise Reasoning Accuracy Reward (StepRAR) and Step-wise Reasoning Validity Reward (StepRVR).

StepRAR rewards the reasoning path using a soft key-step matching technique that evaluates whether the reasoning path contains key intermediate reasoning steps (i.e., the necessary steps to reach the correct final solution).
StepRVR rewards the reasoning path based on a reasoning completeness and logic evaluation method, which assesses whether the reasoning process is well-structured and logically consistent.
In this way, StepRAR and StepRVR help mitigate the sparse reward issue by providing informative rewards, even when the reasoning path does not produce the correct final answer -- as long as it includes key intermediate reasoning steps or follows a structured and logical reasoning process.
With StepRAR and StepRVR, StepGRPO takes the average step-wise reasoning rewards of a group of sampled reasoning paths as a baseline to estimate the advantage for policy optimization. 
Using the proposed StepGRPO, we develop R1-VL, a series of MLLMs with R1-like step-by-step reasoning capabilities.

The proposed StepGRPO offers two key advantages. 
1) Effectiveness. StepGRPO introduces two step-wise reasoning reward mechanisms with group relative optimization, which provide rich and fine-grained step-wise reasoning rewards along the whole reasoning trajectory beyond the final answer. This mitigates the sparse reward issue and encourages more structured, logically consistent reasoning trajectories.
2) Efficiency. StepGRPO achieves step-wise reasoning rewarding in a rule-based manner, which provides step-wise reasoning rewards while eliminating the need of process reward models. 
This significantly reduces computational overhead while maintaining fine-grained step-wise supervisions.

The main contributions of this work are threefold. 
First, we propose StepGRPO, a new online reinforcement learning framework that enables MLLMs to self-improve reasoning ability via a simple, effective and dense step-wise rewarding. 
Second, we design two novel rule-based reasoning reward mechanisms, i.e., step-wise reasoning accuracy reward and step-wise reasoning validity reward, which effectively mitigate the sparse reward issue for MLLMs without the need of process reward models.
Third, with the proposed StepGRPO, we develop R1-VL, a series MLLMs that have superior reasoning capabilities.
Forth, extensive experiments over multiple benchmarks show that R1-VL achieves superior performance compared with state-of-the-art MLLMs.

\section{Related Work}
\label{related_works}

\subsection{Multimodal Large Language Model}

Multimodal Large Language Models (MLLMs)~\cite{gpt4o,claude_3.5_sonnet,Cambrian-1,mm1.5,idefics3,internvl2,deepseek-vl2,llava,vision_survey} have shown remarkable advancements across a wide range of vision-language understanding tasks, demonstrating their capabilities in comprehending and analyzing visual contents across various application domains. Early research on MLLMs primally focuses on text generation based on text prompts and input multiple modalities such as images~\cite{llava,llavanext,zhang2024historical}, videos~\cite{sun2024video,cheng2024videollama}. Recent advancements further enhance the capabilities of MLLMs from various aspects. For example, recent models~\cite{lyu2023macaw,wu2023next} incorporate multimodal inputs and outputs such as video, audio, and point cloud inputs beyond text and images. In addition, some efforts attempt to adapt MLLMs for domain-specific tasks, such as medical image understanding~\cite{zhang2023pmc,li2023llavamed,lan2025gem} and document analysis~\cite{ye2023mplugdoc,liu2024textmonkey}.
In this work, we focus on enhancing the reasoning ability of MLLMs in tackling complex reasoning tasks and introduce R1-VL, a series of MLLMs that have superior reasoning capability.

\subsection{MLLM Reasoning} 

Inspired by the advances in NLP that show great potential in learning to reason and tackling complex language tasks~\cite{openai2024o1}, recent studies attempt to enhance the reasoning capability of MLLM. Generally, current MLLM reasoning methods improve the reasoning capability of MLLM by generating high-quality chain-of-thoughts (CoT) data using powerful model (e.g., GPT-4) and performing supervised fine-tuning with the collected data~\cite{llava-cot,llava-reasoner,thawakar2025llamav,yao2024mulberry,insight-v}.
For example, Mulberry~\cite{yao2024mulberry} introduces Collective Monte Carlo Tree Search (MCTS) into MLLM and proposes CoMCTS which leverages complementary knowledge from multiple models to collaboratively search and identify effective reasoning paths. 
In addition, recent works~\cite{yao2025r1,peng2025lmm,meng2025mm,huang2025vision} attempt to explore online reinforcement learning to improve the MLLMs’ reasoning ability.
Different from these works, we design StepGRPO that enables MLLM to self-improve the reasoning ability with step-wise reward signals.

\subsection{Reinforcement Learning}

Reinforcement Learning (RL)~\cite{kaelbling1996reinforcement} is a fundamental approach in machine learning, where an agent learns to interact with an environment by taking actions, receiving rewards, and updating its policy to maximize the long-term return. 
With the rise of large language models (LLMs)~\cite{radford2018improving,brown2020language,openai2023gpt4}, Reinforcement Learning with Human Feedback (RLHF)~\cite{bai2022training} has emerged as a key technique for fine-tuning models using human preference data. RLHF leverages algorithms like Proximal Policy Optimization (PPO)~\cite{PPO} and Direct Preference Optimization (DPO)~\cite{DPO} to guide model behavior for improving the alignment, coherence and helpfulness in response generation.

Recently, RL is increasingly adopted to enhance LLMs' reasoning capabilities~\cite{zhang2024rest,chen2024step,chen2024self,luong2024reft,guo2025deepseek,team2025kimi}, especially for mathematical problem solving. 
The core is to adopt an appropriate reward function or model that evaluates and reinforces high-quality reasoning paths while penalizing low-quality ones, guiding the model's optimization towards more structured and coherent reasoning trajectories using the RL algorithm.
For example, ReST-MCTS*~\cite{zhang2024rest} trains a process reward model (PRM) for  determining the correctness of each reasoning step within reasoning paths.
Recent methods have found that using a simple outcome-level rule-based reward function (i.e., the reasoning trajectories leading to correct answer are rewarded with higher score) can already provide an effective and reliable reward signal during the RL process~\cite{luong2024reft,guo2025deepseek,team2025kimi}. For example, DeepSeek-R1~\cite{guo2025deepseek} demonstrates that group relative policy optimization (GRPO)~\cite{shao2024deepseekmath} with outcome-level reward effectively enhances the reasoning capability of LLMs.
In this work, we aim for improving the reasoning capability of MLLMs through reinforcement learning and propose StepGRPO, which effectively tackles the sparse reward issue in MLLMs, leading to stable training process and better reasoning capability.

\begin{figure*}[t]
\centering
\includegraphics[width=1\linewidth]{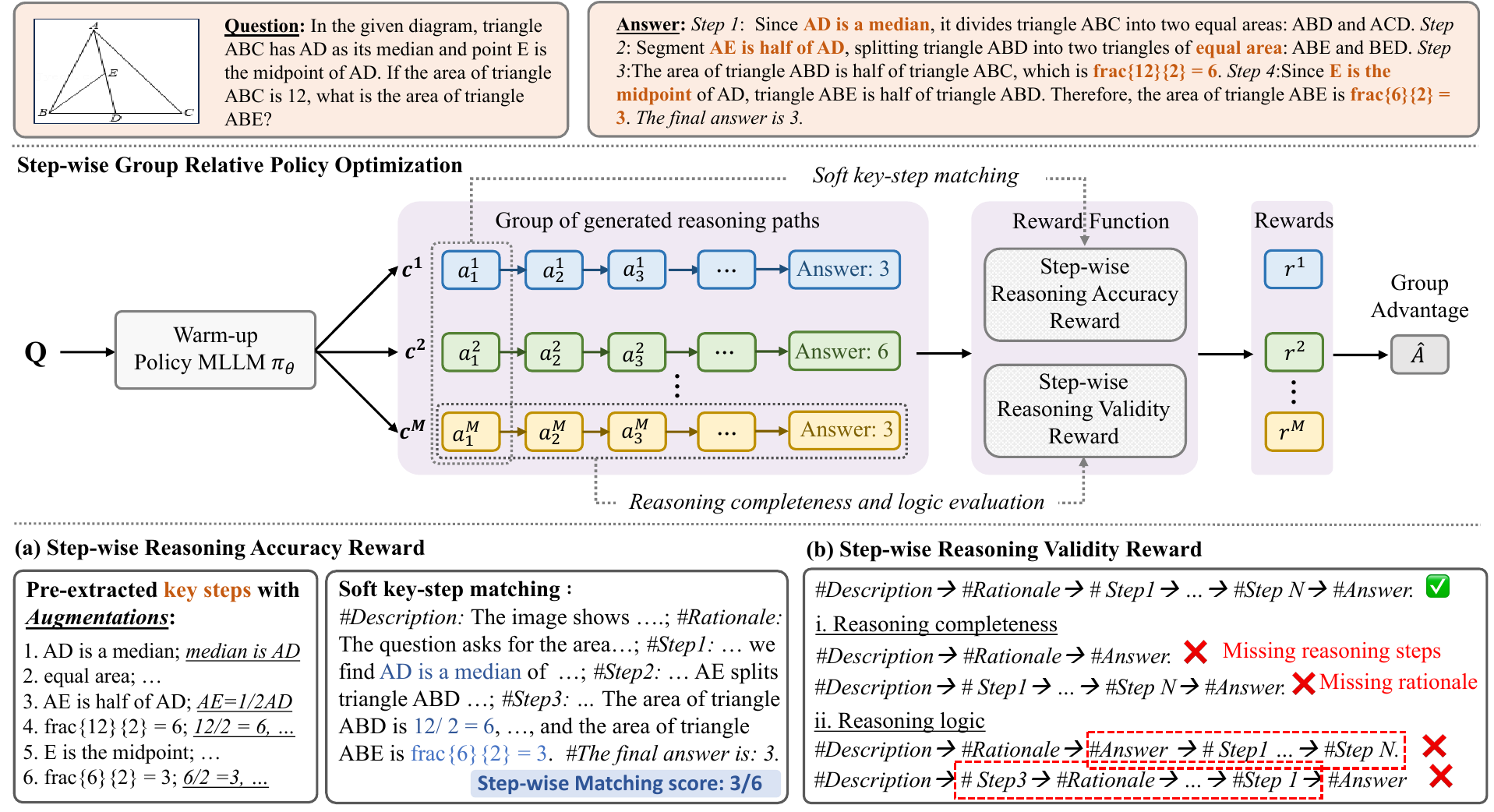}
\caption{
Overview of the proposed StepGRPO. StepGRPO consists of two phases: a policy warm-up phase and a step-wise online policy optimization phase. After the warm-up, the policy model $\pi_{\theta}$ generates a group of reasoning paths $\{\mathbf{c}^i\}_{i=1}^M$ and assigns step-wise rewards using two proposed mechanisms: Step-wise Reasoning Accuracy Reward (StepRAR) and Step-wise Reasoning Validity Reward (StepRVR).
StepRAR rewards reasoning paths that contain key intermediate steps, identified using a soft key-step matching technique. 
StepRVR rewards reasoning paths based on completeness and logical consistency, ensuring they are well-structured. 
StepGRPO then estimates the advantage $\hat{A}$ for policy optimization by using the average step-wise reasoning reward of a group of sampled reasoning paths as a baseline. Examples for StepRAR and StepRVR are illustrated in (a) and (b), respectively.}
\label{fig2}
\end{figure*}

\section{Method}
\label{sec3}

This section first presents the task formulation, and then introduces the proposed Step-wise Group Relative Policy Optimization (StepGRPO). More details to be elaborated in the ensuing subsections.

\subsection{Task Formulation}

In this paper, we consider a pre-trained MLLM and denote it as a policy model $\pi_{\theta}$. 
Given a multimodal question $Q$  consisting of an image and a textual task instruction, i.e., $Q = \{\text{text}, \text{image}\}$, the policy model $\pi$ generates response $\mathbf{c}$ with a step-by-step reasoning trajectory. 
Generally, this process can be formulated as a sequence of next token prediction actions, i.e., $ \mathbf{c} = (a_1, a_2, ..., a_t, ..., a_T)$, where each action $a_t$ is sampled from the policy model $\pi_{\theta}$ and $T$ represents the maximum sequence length.
After each action, the new state $s_{t+1}$ is determined by updating the current state $s_t$ with the newly generated action $a_t$, i.e., $s_{t+1} = (s_t, a_t), 1\leq t \leq T$.

Considering this formulation, the objective of our task is to optimize the policy model $\pi_{\theta}$ such that it can select better actions based on the previous states, thereby improving reasoning quality. In the context of reinforcement learning (RL), the policy model is generally optimized by maximizing the cumulative reward, where the reward for taking action $a_t$ at state $s_t$ is denoted as $r(s_t,a_t,s_{t+1})$. 
Following prior studies~\cite{yao2024mulberry}, we define an action in this paper as generating a reasoning step, which consists of one or more sentences containing multiple word tokens.

\subsection{Step-wise Group Relative Policy Optimization }

We propose Step-wise Group Relative Policy Optimization (StepGRPO), a novel online reinforcement fine-tuning framework that mitigates the sparse reward issue for MLLMs and encourages self-improvement in reasoning ability through simple, effective and dense step-wise reward mechanisms.
As illustrated in Fig.~\ref{fig2}, StepGRPO consists of two phases: (1) a policy warm-up phase and (2) a step-wise online policy optimization phase.
The overall algorithm is shown in Algorithm~\ref{alg}.

\subsubsection{Policy Warm-up}

This phase equips the policy model with fundamental reasoning capabilities, ensuring it can generate proper step-wise reasoning paths before reinforcement learning.
During the warm-up phase, the policy model is fine-tuned using a multimodal dataset $D_{s}$ with Chain-of-Thought (CoT) reasoning path, where each data consists of a multimodal question $Q$ and a step-by-step reasoning path $\tau$, i.e., $D_{s} = \{Q^n, {\tau}^n\}_{n=1}^N$:
\begin{equation}
    \mathcal{L}_{warm-up} = -\mathbb{E}_{\tau \sim D_{s}}[\sum_{t=1}^{T}\text{log}(\pi_{\theta}(a_t|s_t))].
\label{eq1}
\end{equation}

\subsubsection{Step-wise Online Policy Optimization}
This phase enables MLLMs to self-improve their reasoning ability via online reinforcement learning, mitigating the sparse reward issue through step-wise reasoning rewards.
As illustrated in Fig.~\ref{fig2}, for each question $Q \in D_{s}$, the policy model $\pi_{\theta}$ first generates a group of $M$ reasoning trajectories via multiple rollouts, i.e., $\{\mathbf{c}^i\}_{i=1}^M$, where $ \mathbf{c}^i = (a_1^i, a_2^i, ..., a_t^i, ..., a_T^i)$.
After obtaining a group of $M$ reasoning trajectories, we employ our proposed step-wise reasoning rewards to evaluate and reward each generated reasoning trajectory. Specifically, we introduce two types of rule-based step-wise rewards, i.e., step-wise reasoning accuracy (StepRAR) reward and step-wise reasoning validity reward (StepRVR).

\noindent\textbf{Step-wise reasoning accuracy reward (StepRAR)} reduces the effect of learning from sparse reward by additionally rewarding reasoning paths that contain correct intermediate reasoning steps contributing to the final solution. 
Specifically, for each question $Q$, we pre-extract a set of key reasoning steps $\mathbf{v} = \{v_1, v_2, ...\}$ from the corresponding reasoning path $\tau$ in dataset $D_{s}$.
We define key steps as the essential variables and equations that directly contribute to the final solution, and prompt GPT-4 to extract several key steps from the reasoning path for each question.
To ensure efficient reward assignment, we refine the extracted steps by removing redundant content and retaining only the core few words necessary for reasoning. Furthermore, we augment each extracted key step into multiple equivalent formats to allow more flexible and accurate matching, preventing missed matches due to math-related formatting differences. For example, a mathematical expression such as ``$\backslash$frac\{6\}\{3\} = 2'' is augmented to ``6/3 = 2'' or ``6 divided by 3 equals 2''.

With the extracted key reasoning steps $\mathbf{v} = \{v_1, v_2, ...\}$ and such soft marching mechanism, we calculate a match score for each generated reasoning path based on the ratio of matched key steps, i.e., $k^i = {|\mathbf{v}_{match}|} / {|\mathbf{v}|}$. 
Then, StepRAR for $1 \leq t \leq T$ is defined as: 
\begin{equation}
r_{auc}^i(s_t, a_t, s_{t+1}) =
\begin{cases} 
    1 + \alpha k^i, & \text{ans}(s_{t+1}) = y, \\
    \alpha k^i , & \text{ans}(s_{t+1}) \neq \text{null}, \neq y, \\
    0, & \text{ans}(s_{t+1}) = \text{null},
\end{cases}
\label{eq2}
\end{equation}
where $y$ is the ground-truth answer extracted from CoT reasoning path.

By leveraging pre-extracted key reasoning steps, StepRAR efficiently provides additional supervision with a simple soft matching mechanism, ensuring the model learns meaningful reasoning processes instead of guessing answers randomly.

\noindent\textbf{Step-wise reasoning validity reward (StepRVR)} aims for ensuring
the generated paths adhere to a logically structured and coherent progression beyond the reasoning accuracy. Prior studies~\cite{llava-cot,yao2024mulberry} have demonstrated structural reasoning, such as problem decomposition and progressive reasoning, facilitates more accurate and interpretable reasoning processes, as they encourage models to break down complex problems into multiple intermediate steps rather than direct answer generation.

Inspired by these findings, we incorporate step-wise reasoning validity to reinforce well-organized reasoning paths that follow an expected logical flow. Specifically, we define StepRVR using two key criteria: reasoning completeness $\delta^c$ and reasoning logic $\delta^l$. Reasoning completeness requires the response to include three essential components, i.e., a background analysis involving image description and rationale analysis to establish context, a step-by-step reasoning process and a final answer. In addition to the reasoning completeness, reasoning logic ensures the reasoning path to follow a logical progression, where the background analysis must come before solution steps and the final answer should only appear after reasoning steps are complete.

With these two criteria, we define StepRVR as 
\begin{equation}
r_{val}^i(s_t, a_t, s_{t+1}) =
\begin{cases} 
    1, & \mathbb{I}(\delta^c(s_{t+1}))\cdot\mathbb{I}(\delta^l(s_{t+1})) = 1, \\
    0, & \text{otherwise},
\end{cases}
\label{eq3}
\end{equation}
where the reasoning trajectory is rewarded only if it satisfies both completeness and logical coherence. By enforcing this, StepRVR helps the model produce structured, interpretable and logically sound reasoning trajectories, enhancing both the quality and reliability of generated responses.

\begin{algorithm}[t]
\caption{Step-wise Group Relative Policy Optimization}
\label{alg}
\begin{algorithmic}
    \STATE {\bfseries Input:} Policy model $\pi_{\theta}$ initialized by a pre-trained MLLM; a multimodal dataset $D_{s} = \{Q^n,{\tau}^n\}_{n=1}^N$.
    \STATE {\bfseries Output:} Trained policy model $\pi_{\theta}$
    \STATE \textcolor{gray}{Policy warm-up:}
    \FOR{\textit{iter = 1 to $N$}}
    \STATE Sample $\{Q,{\tau}\} \in D_{s}$ 
    \STATE Optimize policy model $\pi_{\theta}$ by Eq.~\ref{eq1}
    \ENDFOR
    \STATE \textcolor{gray}{ Step-wise online policy optimization:}
    \FOR{\textit{iter = 1 to $N$}}
    \STATE Sample $\{Q,{\tau}\} \in D_{s}$
    \STATE Generate a group of reasoning paths $\{\mathbf{c}^i\}^M_{i=1} \sim \pi_{\theta}$
    \STATE Obtain step-wise rewards $\{r^i\}_{i=1}^M$ by Eqs.~\ref{eq2}-\ref{eq3}
    \STATE Obtain relative advantages $\{\hat{A}^i\}_{i=1}^M$ by Eq.~\ref{eq4} 
    \STATE Optimize policy model $\pi_{\theta}$ by Eqs.~\ref{eq5}-\ref{eq6} 
    \ENDFOR
  \RETURN policy model $\pi_{\theta}$
\end{algorithmic}
\end{algorithm}

\noindent\textbf{Optimization with the step-wise rewards.}
After obtaining two types of step-wise rewards, we compute the overall reward for each reasoning path as $r^i = r_{auc}^i + r_{val}^i$, and repeatedly compute the rewards for all generated reasoning paths, i.e., $\{r^1, r^2, ..., r^M\}$. 

To estimate the advantage of each reasoning trajectory, we normalize its reward relative to the group as follow:
\begin{equation}
    \hat{A}^i = \frac{r^i - \text{mean}(\{r^1, r^2, ..., r^M\})}{\text{std}(\{r^1, r^2, ..., r^M\})},
\label{eq4}
\end{equation}
where the mean group reward serves as the baseline, and $\hat{A}_i$ measures how much better or worse $r_i$ is compared to other reasoning trajectories within the group.
Following this, we optimize the policy model with the loss defined as:
\begin{equation}
\begin{split}
    \mathcal{L}_{StepRL} & = -\underset{Q\in D_s}{\mathbb{E}} [\frac{1}{M}\sum_{i=1}^M (\frac{\pi_{\theta}(\mathbf{c}^i|Q)}{[\pi_{\theta}(\mathbf{c}^i|Q)]_{\text{no grad}}}\hat{A}^i  \\ &- \beta D_{KL}(\pi_{\theta}||\pi_{ref})],
\end{split}
\label{eq5}
\end{equation}
where KL divergence is adopted to regularize the policy model, preventing excessive deviation from the reference model. The reference model is typically initialized as the same model as the policy model but remains frozen during RL training. The KL divergence between the policy model and the reference model is estimated as in~\cite{shao2024deepseekmath}:
\begin{equation}
D_{KL}(\pi_{\theta}||\pi_{ref}) = \frac{\pi_{ref}(\mathbf{c}^i|Q)}{\pi_{\theta}(\mathbf{c}^i|Q)} - \text{log} \frac{\pi_{ref}(\mathbf{c}^i|Q)}{\pi_{\theta}(\mathbf{c}^i|Q)} -1.
\label{eq6}
\end{equation}

\section{Experiment}

This section presents experiments including datasets and implementation details, main experimental results, ablation studies and discussion, respectively. More details are to be described in the ensuing subsections.

\begin{table*}[ht]
  \centering
  \scalebox{0.9}{
  \setlength{\tabcolsep}{2pt}
  \begin{tabular}{@{}lcccccccccl@{}}
    \toprule
    Method & MathVista & MMStar  & Math-V  & ChartQA & DynaMath & HallBench & MathVerse & MME$_{sum}$ & MMReason & \makecell[c]{AVG} \\
    \midrule
    \textit{Closed-Source Model} \\
    GPT-4o~\cite{gpt4o} & 63.8 & 63.9  & 30.3& 85.7 & 63.7 & 55.0 & 39.4 & 2329 & 21.1&56.2 \\
    Claude-3.5 Sonnet~\cite{claude_3.5_sonnet}  & 67.7 & 62.2  &-& 90.8 & 64.8 & 55.0 & - & 1920 & \makecell[c]{-} &\makecell[c]{-} \\
    \midrule
    \textit{Open-Source Model} & & & &\\
    Cambrain-1-8B~\cite{Cambrian-1} & 49.0 & - &- & 73.3 & - & - & - & - &  \makecell[c]{-} & \makecell[c]{-}\\ 
    MM-1.5-7B~\cite{mm1.5} & 47.6 & - &-& 78.6 & - & - & - & 1861 & \makecell[c]{-} &\makecell[c]{-} \\ 
    Idefics3-LLaMA3-8B~\cite{idefics3} & 58.4 & 55.9 & -& 74.8 & - & - & - & 1937 & \makecell[c]{-}&\makecell[c]{-}\\ 
    InternVL2-8B~\cite{internvl2} & 58.3 & 61.5 & -& 83.3 & 39.7 & - & - & 2210 & \makecell[c]{-}&\makecell[c]{-}\\
    MiniCPM-V-2.6-8B~\cite{minicpm-v} & 60.6 & 57.5 & -& - & - & 48.1 & - & 2348 & \makecell[c]{-} &\makecell[c]{-}\\
    DeepSeek-VL2-MOE-4.5B~\cite{deepseek-vl2} & 62.8 & 61.3 &- & 86.0 & - & - & - & 2253 & 11.5&\makecell[c]{-} \\
    \midrule
    \textit{Reasoning Model} & & & &\\ 
    LLaVA-CoT-11B~\cite{llava-cot} & 54.8 & 57.6 & - & -& - & 47.8 & - & - & \makecell[c]{-}&\makecell[c]{-}\\ 
    LLaVA-Reasoner-8B~\cite{llava-reasoner} & 50.6 & 54.0 & -& 83.0 & - & - & - & - & \makecell[c]{-} &\makecell[c]{-}\\ 
    Insight-V-8B~\cite{insight-v} & 49.8 & 57.4  &-& 77.4 & - & - & - & 2069 & \makecell[c]{-}&\makecell[c]{-}\\
    Mulberry-7B~\cite{yao2024mulberry}  & {63.1} & 61.3  &-& {83.9} & {45.1} & {54.1} &- & {2396}  &11.8&\makecell[c]{-}\\
    LlamaV-o1-11B~\cite{thawakar2025llamav}& 54.4& 59.4 & -& -&-&63.5&-&-&-&\makecell[c]{-} \\
    Vision-R1-7B~\cite{huang2025vision} &73.5&-&-&-&-&-&52.4 &-&-&\makecell[c]{-} \\
    LMM-R1~\cite{peng2025lmm}&63.2&58.0&26.3&-&-&-&41.5&-&-& \makecell[c]{-}\\
    R1-ShareVL-7B~\cite{yao2025r1}&75.4&67.0&29.5&-&-&-&52.8&-&-&\makecell[c]{-} \\
    \midrule 
    Qwen2-VL-2B~\cite{qwen2vl} & 43.0 & 48.0  &12.4 &73.5 & 24.9 & 41.7 & 19.7  & 1872  &7.7&37.5 \\
    \rowcolor{mygray}
    \textbf{R1-VL-2B (Ours)} & 52.1 & 49.8&17.1&75.2&29.4&44.0&26.2&2048&8.3 &41.6 \\
    \midrule
    Qwen2-VL-7B~\cite{qwen2vl} & 58.2 & 60.7  &16.3& 83.0 & 42.1 & 50.6 & 32.5 &2327 &11.9 &48.7 \\
    \rowcolor{mygray}
    \textbf{R1-VL-7B (Ours)} & {63.5} & 60.0&24.7 & 83.9 &45.2&54.7&40.0&2376&12.5&52.1 \\
    \midrule
    Qwen2.5-VL-7B~\cite{bai2025qwen2_5} & 68.2&63.9&25.1&87.3&53.2&52.1&49.2&2347&17.3&55.5 \\
    \rowcolor{mygray}
    \textbf{R1-VL-7B* (Ours)} &74.3&66.2&28.2&87.7&56.5&57.2&52.2&2395&17.9& 58.4\\
    \bottomrule
  \end{tabular}}
  \caption{Main experimental results. To comprehensively examine the proposed StepGRPO, we conduct extensive experiments with two baseline models on eight benchmarks, and compare StepGRPO with various state-of-the-art MLLMs.* indicates that the model is trained using Qwen2.5-VL-7B as the base model with the data from~\cite{yao2025r1}.
  }
  \label{tab:SOTA}
\end{table*}

\begin{table}[t]
  \centering
  \resizebox{0.99\linewidth}{!}{
  \setlength{\tabcolsep}{3pt}
  \begin{tabular}{@{}c|cc|c@{}}
    \toprule
    \multirow{2.5}{*}{ \quad Warm-up \quad } &\multicolumn{2}{c|}{\quad Step-wise reasoning rewards \quad } & \multirow{2.5}{*}{\quad MathVista \quad } \\
    \cmidrule(){2-3}
     & \ \quad StepRAR \quad \ & \quad \ StepRVR \ \quad &  \\
    \midrule
     & & & 58.2 \\
    \Checkmark & & & 61.2 \\
    \Checkmark &\Checkmark & & 62.4 \\
    \Checkmark & &\Checkmark & 61.9 \\
    \Checkmark & \Checkmark &\Checkmark & \textbf{63.5} \\
    \bottomrule
  \end{tabular}}
  \caption{
  Ablation study of StepGRPO over Qwen2-VL-7B. 
  }
  \label{tab:Ablation}
\end{table}

\subsection{Datasets}

For policy warm-up, we adopt Mulberry-260k~\cite{yao2024mulberry} for supervised fine-tuning. For step-wise online policy optimization, we randomly sample 10K data from Mulberry-260k as our training data.
For evaluation, we adopt 8 widely-used multimodal benchmarks for comprehensively evaluating our proposed StepGRPO, including MathVista~\cite{mathvista}, MMStar~\cite{mmstar}, Math-Vision~\cite{wang2025measuring}, ChartQA~\cite{masry2022chartqa}, DynaMath~\cite{dynamath}, HallusionBench~\cite{hallusionbench}, MathVerse~\cite{zhang2024mathverse}, MME~\cite{fu2023mme} and MMReason~\cite{yao2025mmreason}. These multimodal benchmarks cover a wide range of tasks from mathematical reasoning, chart understanding, visual hallucination and general visual understanding.

\subsection{Implementation Details}
\label{implement}

Our proposed StepGRPO is generally applicable to different MLLMs. In our experiments, we adopt two state-of-the-art open-source MLLMs, i.e., Qwen2-VL-2B and Qwen2-VL-7B~\cite{qwen2vl}. 
For the policy warm-up phase, we set the training batch size to 128. Following prior work~\cite{yao2024mulberry}, we use a learning rate of $1\text{e}^{-5}$ for Qwen2-VL-2B and $5\text{e}^{-6}$ for Qwen2-VL-7B, respectively.

For the step-wise online policy optimization phase, we perform 4 rollouts per question ($M = 4$) and set the sampling temperature to 1.2 to encourage diverse reasoning paths. The maximum sequence length is set to $L = 1024$, ensuring that the model can generate complete reasoning paths. Both the policy model and reference model are initialized from the model after the warm-up, with the reference model frozen during RL training. The policy model’s learning rate is $1\text{e}^{-6}$, and we set the batch size to 4. We set the coefficient of match score $\alpha$ to 0.1 to balance its effect. Following~\cite{vonwerra2022trl}, the KL divergence coefficient $\beta$ in Eq.~\ref{eq5} is set to 0.04 by default.
All experiments are conducted on 4 H100-80GB GPUs.

\subsection{Main Experimental Results}

We conduct a comprehensive evaluation of R1-VL across eight widely used benchmarks, comparing it with various state-of-the-art MLLMs, as shown in Table~\ref{tab:SOTA}.

We first compare R1-VL with its baseline models, Qwen2-VL-2B and Qwen2-VL-7B. 
The baseline models exhibit limited reasoning capability, leading to very few reasoning paths receiving rewards, which negatively impacts the reasoning capability. In contrast, R1-VL with our proposed StepGRPO consistently improves the baseline models by significant margins, achieving 4.6\% improvement over Qwen2-VL-2B and 3.8\% over Qwen2-VL-7B. This improvement is largely attributed to that StepGRPO introduces step-wise reasoning accuracy and validity rewards, which provide rich and informative supervision at each reasoning step, effectively mitigating the sparse reward issue for MLLMs.

In addition, we compare R1-VL with existing state-of-the-art reasoning MLLMs. As shown in Table~\ref{tab:SOTA}, R1-VL achieves better performance on most benchmarks, particularly in mathematical reasoning tasks. For example, R1-VL-7B surpasses Mulberry-7B and LlamaV-o1-11B by 0.6\% and 9.3\% respectively on the reasoning-intensive benchmark MathVista. Notably, R1-VL-2B even outperforms larger MLLMs. For instance, R1-VL-2B largely outperforms LLaVA-Reasoner-8B and LLaVA-CoT-11B by 13.1\% and 9.3\% on MathVista, respectively. This superior performance demonstrates that StepGRPO effectively enhances MLLMs’ reasoning abilities by encouraging self-improvement via step-wise online reinforcement learning, rather than merely imitating positive reasoning paths.

Additionally, we benchmark R1-VL against general MLLMs, including closed-source models such as GPT-4o and Claude-3.5 Sonnet, as well as open-source models like Cambrain-1-8B and DeepSeek-VL2-MOE-4.5B. We observe that R1-VL outperforms most open-source MLLMs and achieves competitive results against closed-source models. For example, R1-VL-7B achieves 63.7 accuracy on MathVista, closely matching GPT-4o's accuracy of 63.8. These results further validate StepGRPO’s effectiveness in enhancing the reasoning capabilities of MLLMs.

\begin{table}[t]
\centering
\begin{footnotesize}
\resizebox{1.0\linewidth}{!}{
\begin{tabular}{c|ccccc}
 \toprule
 &\multicolumn{5}{c}{Number of generations $M$ per question} \\
 \midrule
 Method  & \quad 2   \quad & \quad 3  \quad & \quad 4  \quad & \quad 5  \quad & \quad 6  \quad \\
 \midrule
\ \ \ \quad R1-VL-7B \ \ \ \ \quad  & 62.5 & 62.8 & {63.5} & 63.2 &63.7 \\
\bottomrule
\end{tabular}
}
\end{footnotesize}
\caption{Parameter analysis of $M$. The experiments are conducted on Qwen2-VL-7B over MathVista.}
\label{dis:para_G}
\end{table}

\subsection{Ablation Study}

We conduct ablation studies for StepGRPO on Qwen2-VL-7B over MathVista benchmark for examining the effect of step-wise reasoning rewards including step-wise reasoning accuracy reward (StepRAR) and step-wise reasoning validity reward (StepRVR), as well as the role of the warm-up phase. 
As shown in Table~\ref{tab:Ablation}, involving a warm-up stage improves baseline model to 61.2\%, allowing the model to learn basic reasoning knowledge before reinforcement learning.
In addition, including either StepRAR or StepRVR into the online reinforcement learning process outperforms the model with warm-up by large margins, demonstrating that both two types of step-wise rewards contribute to enhancing step-by-step reasoning capabilities.
The best performance (i.e., 63.7\%) is achieved when both StepRAR and StepRVR are applied together. 
This shows that StepGRPO effectively improves complex reasoning tasks by reinforcing both the correctness of intermediate steps and the overall logical structure of the reasoning process.

\begin{table}[t]
\centering
\begin{footnotesize}
\resizebox{\linewidth}{!}{
\begin{tabular}{l|c}
 \toprule
 Method& \ \ MathVista \ \ \\
 \midrule
Warm-up \quad & 61.7 \\
Warm-up + Outcome-level reward& 62.3 \\
\textbf{Warm-up + Step-wise reward (Ours)} \quad \quad & \textbf{63.5} \\
\bottomrule
\end{tabular}
}
\end{footnotesize}
\caption{Effectiveness of the step-wise reasoning rewards. The experiments are conducted on Qwen2-VL-7B over MathVista.}
\label{dis:step}
\end{table}

\begin{figure}[t]
\centering
\includegraphics[width=0.9\linewidth]{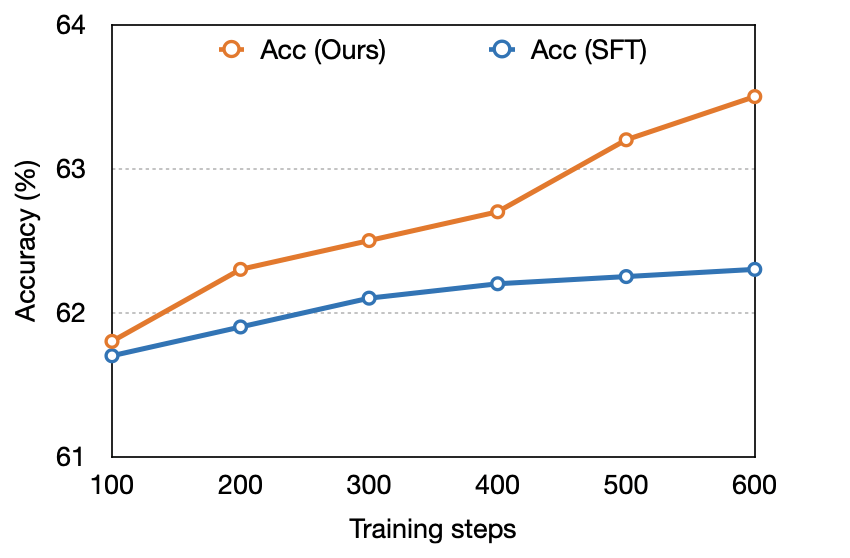}
\caption{Comparison between StepGRPO and SFT. The experiments are conducted on Qwen2-VL-7B over MathVista.
}
\label{fig:3}
\end{figure}

\begin{figure*}[!t]
\centering
\includegraphics[width=0.99\linewidth]{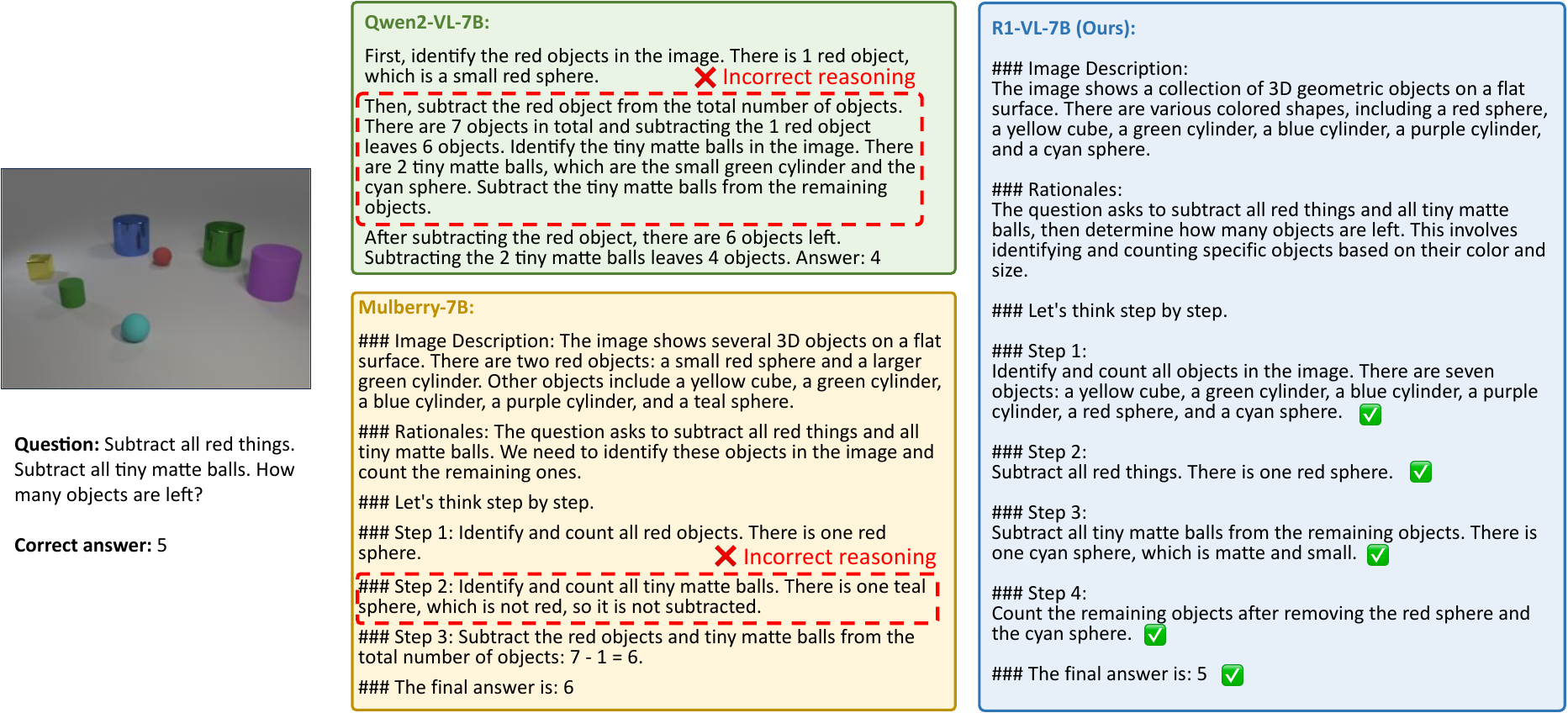}
\caption{Qualitative comparison. }
\label{fig4}
\end{figure*}

\subsection{Discussion}
\label{Discuss}

\noindent \textbf{Parameter analysis.}
We conduct the parameter analysis on the number of generations $M$ in a group with Qwen2-VL-7B over benchmark MathVista, analyzing its impact on reasoning performance.
As described in Section~\ref{sec3}, $M$ controls the number of generated reasoning trajectories per question during the RL phase. Table~\ref{dis:para_G} shows that a larger $M$ generally leads to better performance. This is because, in group relative optimization, the baseline reward is estimated as the average reward of all generated reasoning paths. A larger $M$ results in a more stable and accurate baseline estimation, whereas a small $M$ may lead to high variance in baseline estimation, making the optimization process less reliable.
However, increasing $M$ also introduces higher computational costs. Therefore, we set $M=4$ as the default to balance performance and computational efficiency.

\noindent\textbf{Effectiveness of the step-wise reward.} 
Our proposed step-wise reward mechanism plays a crucial role in mitigating the sparse reward issue by providing fine-grained supervision at each reasoning step. To further validate its effectiveness, we conduct an experiment comparing outcome-level reward against our step-wise reward.
Specifically, we evaluate three settings: (1) Warm-up only ; (2) Warm-up + Outcome-level Reward, where the model is optimized with outcome-level reward; and (3) Warm-up + Step-wise Reward, where the model is optimized with our proposed step-wise reasoning reward.
As shown in Table~\ref{dis:step}, both outcome-level reward and our step-wise reward improve the warm-up model’s performance, while our step-wise reward achieves better performance. This further demonstrates that step-wise rewards are more effective in enhancing MLLMs’ reasoning capabilities, as they provide more fine-grained supervision and largely mitigate the sparse reward issue.

\noindent\textbf{Comparison to supervised fine-tuning (SFT).}
As discussed before, StepGRPO encourages MLLM to self-improve the reasoning ability with step-wise reward signals rather than merely imitating the successful reasoning paths.
Here, we conduct experiments to further compare StepGRPO with SFT. Specifically, we start with the model after the warm-up and conduct the experiments with Qwen2-VL-7B over MathVista.
As shown in Fig.~\ref{fig:3}, under the same number of training steps, StepGRPO consistently outperforms SFT, demonstrating the effectiveness of step-wise reinforcement learning. This is largely attributed to StepGRPO’s ability to refine reasoning trajectories through self-exploration and reward-guided optimization, rather than solely relying on passive imitation of reasoning paths. By leveraging step-wise reasoning rewards, StepGRPO provides more rich and informative supervision, leading to better reasoning processes compared to SFT.

\noindent \textbf{Qualitative comparison.} We provide qualitative comparison of Qwen2VL-7B, Mulberry-7B and our R1-VL-7B. As shown in Fig.~\ref{fig4}, Qwen2-VL-7B generates relatively short responses, lacking a thorough reasoning process. While Mulberry-7B generates detailed reasoning paths, its intermediate steps contain errors, leading to incorrect final answers. In contrast, R1-VL-7B enables more accurate step-by-step reasoning process.

We provide more discussions, experimental results and
qualitative analysis in the appendix.

\section{Conclusion}

This paper presents StepGRPO, a new online reinforcement learning framework that enables MLLMs to self-improve reasoning ability via simple, effective and dense step-wise reward mechanism. Specifically, StepGRPO introduces two rule-based reasoning reward mechanisms, i.e., Step-wise Reasoning Accuracy Reward that rewards the intermediate reasoning steps based on a soft key-step matching technique and Step-wise Reasoning Validity Reward that rewards the reasoning path's reasoning structure and logical consistency though a reasoning completeness and logic evaluation method. 
In this way, StepGRPO enables to effectively mitigate the sparse reward issue for MLLMs without the need of process reward models and encourages more structured and logically consistent reasoning process. With the proposed StepGRPO, we develop R1-VL, a series of MLLMs with superior reasoning capability. Extensive experiments over eight benchmarks demonstrate the superiority of the proposed StepGRPO compared with the state-of-the-art MLLMs. 

\textbf{Acknowledgement.} This research is supported by the RIE2025 Industry Alignment Fund – Industry Collaboration Projects (IAF-ICP) (Award I2301E0026), administered by A*STAR, as well as supported by Alibaba Group and NTU Singapore through Alibaba-NTU Global e-Sustainability CorpLab (ANGEL).

{\small
\bibliographystyle{ieee_fullname}
\bibliography{egbib}
}

\end{document}